\pdfoutput=1

\documentclass[11pt]{article}

\usepackage{acl}

\usepackage{times}
\usepackage{latexsym}
\usepackage{graphicx}

\usepackage[T1]{fontenc}

\usepackage[utf8]{inputenc}

\usepackage{microtype}

\usepackage{multirow,tabularx}
\usepackage{array}
\usepackage{here}
\usepackage{float}
\usepackage{multicol}
\usepackage{arydshln}

\def\FigSpaceTop{-2.0ex}
\def\FigSpaceBottom{-2.0ex}

\newcommand{\woft}{{\footnotesize w/o\hspace{0.1em}}FT}

\title{Comparison and Combination of Sentence Embeddings\\
Derived from Different Supervision Signals}

\author{
  Hayato Tsukagoshi \hspace{6ex} Ryohei Sasano \hspace{6ex} Koichi Takeda \\
  Graduate School of Informatics, Nagoya University\\
  \texttt{tsukagoshi.hayato.r2@s.mail.nagoya-u.ac.jp}, \\
  \texttt{\{sasano,takedasu\}@i.nagoya-u.ac.jp} \\
}

\begin{document}
\maketitle

\begin{abstract}
There have been many successful applications of sentence embedding methods.
However, it has not been well understood what properties are captured in the resulting sentence embeddings depending on the supervision signals.
In this paper, we focus on two types of sentence embedding methods with similar architectures and tasks: one fine-tunes pre-trained language models on the natural language inference task, and the other fine-tunes pre-trained language models on word prediction task from its definition sentence, and investigate their properties.
Specifically, we compare their performances on semantic textual similarity (STS) tasks using STS datasets partitioned from two perspectives: 1) sentence source and 2) superficial similarity of the sentence pairs, and compare their performances on the downstream and probing tasks.
Furthermore, we attempt to combine the two methods and demonstrate that combining the two methods yields substantially better performance than the respective methods on unsupervised STS tasks and downstream tasks.
\end{abstract}

\section{Introduction}
Sentence embeddings are dense vector representations of a sentence.
A variety of methods have been proposed to derive sentence embeddings, including those based on unsupervised learning \cite{SkipThought, FastSent, QuickThoughts, USE, TSDAE} and supervised learning \cite{InferSent}.
Pre-trained Transformer-based \citep{Transformer} language models, such as BERT \cite{BERT} and RoBERTa \cite{RoBERTa}, have been successfully applied in a wide range of NLP tasks, and sentence embedding methods that leverage pre-trained language models have also performed well on semantic textual similarity (STS) tasks and several downstream tasks.
These methods refine pre-trained language models for sophisticated sentence embeddings by unsupervised learning \cite{BERT-flow, SBERT-WK, DeCLUTR, BERT-CT, ConSERT, SimCSE}, or supervised learning \cite{SBERT, DefSent, SimCSE}.

\begin{figure}[t]
\centering
\includegraphics[width=\linewidth]{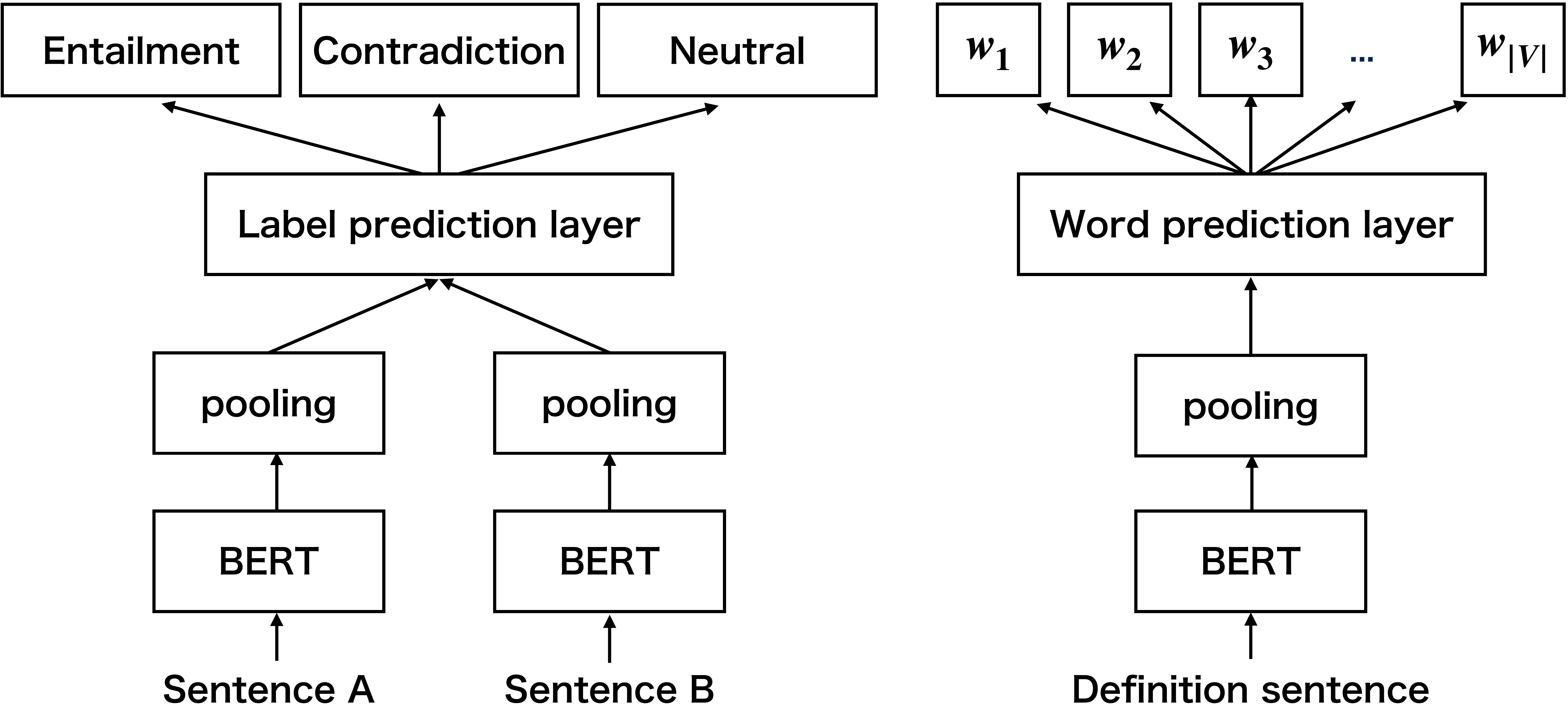}
\vspace{\FigSpaceTop}
\caption{Overviews of SBERT {\small (left)} and DefSent {\small (right)}.}
\label{fig:overview}
\vspace{\FigSpaceBottom}
\end{figure}

Among them, \citet{SBERT} proposed Sentence-BERT (SBERT), which fine-tunes pre-trained language models on the natural language inference (NLI) task.
SBERT performed well on the STS and downstream tasks.
Recently, \citet{DefSent} proposed DefSent, which fine-tunes pre-trained language models on the task of predicting a word from its definition sentence in a dictionary, and reported that it performed comparably to SBERT.
Figure \ref{fig:overview} shows overviews of SBERT and DefSent. 
Although both methods fine-tune the same pre-trained models and use the same pooling operations to derive a sentence embedding, the supervision signals for fine-tuning are different.
That is, SBERT leverages NLI datasets, whereas DefSent leverages word dictionaries.

It is expected that the properties of the sentence embeddings depend on their supervision signals.
However, since existing research has mainly focused on achieving better performance on benchmark tasks, it has not been revealed what property differences the resulting sentence embeddings have. 
Investigating the properties of sentence embeddings would give us a better understanding of existing sentence embedding methods and help develop further methods.
In this paper, we empirically investigate the influence of supervision signals on sentence embeddings.
We focus on SBERT and DefSent because they leverage different supervision signals but have very similar architectures, as shown in Figure \ref{fig:overview};
thus, they would be appropriate for analyzing the influence of the supervision signals on sentence embeddings.

First, we partitioned the STS datasets \citep{STS12, STS13, STS14, STS15, STS16, STSB, SICK} on the basis of two different perspectives and examine what type of meaning each type of sentence embeddings captures by analyzing the performance of each method on these partitioned STS datasets. 
We then apply each type of embeddings to the downstream and probing tasks of SentEval \citep{SentEval} and analyze what type of information is captured.
Our results demonstrate that the supervision signals have a significant impact on performance on these tasks and that the properties of SBERT and DefSent would be complementary.
Thus, we further explore whether combining the two methods yields better sentence embeddings to confirm their complementarity, and demonstrate that combining the two methods yields substantially better performance than the respective methods on unsupervised STS tasks and downstream tasks of SentEval.

\section{Preparation}
In this section, we present detailed descriptions of SBERT and DefSent, the two sentence embedding methods compared in this study, and describe the tasks and settings for the experiments.

\subsection{Sentence-BERT}
\label{SEC::2-1}
Sentence-BERT (SBERT) proposed by \citet{SBERT} is a sentence embedding method that fine-tunes pre-trained language models in a Siamese network architecture on the NLI task.
An overview of SBERT is given on the left side of Figure \ref{fig:overview}\footnote{
Actually, it is possible to use RoBERTa and others instead of BERT, but for simplicity we refer to it as BERT here.
}.
For fine-tuning of SBERT, NLI datasets, such as the Stanford NLI (SNLI) dataset \citep{SNLI} and Multi-Genre NLI (MultiNLI) dataset \citep{MultiGenreNLI}, are used.
These datasets consist of sentence pairs labeled as either entailment, contradiction, or neutral.
The NLI task is a classification task to predict these labels.

SBERT first inputs each sentence of a pair into BERT and obtains sentence embeddings from the output contextualized word embeddings by a pooling operation.
SBERT uses three types of pooling strategies: \texttt{CLS}, which uses the embedding of the first token of the input sequence (e.g., the [CLS] token for BERT); \texttt{Mean}, which uses the average of all word embeddings; and \texttt{Max}, which uses the max-over-time of all word embeddings.
Let $u$ and $v$ be the sentence embeddings obtained by such pooling.
SBERT composes a vector $[u; v; |u - v|]$ and inputs it into a three-way softmax classifier to predict the label of the given sentence pair.

\subsection{DefSent}
\label{SEC::2-2}
DefSent proposed by \citet{DefSent} is a sentence embedding method that fine-tunes pre-trained language models on the task of predicting a word from its definition sentence in a dictionary.
An overview of DefSent is given on the right side of Figure \ref{fig:overview}.
As well as SBERT, DefSent first inputs a definition sentence into BERT and obtains the sentence embedding by a pooling operation, which uses \texttt{CLS}, \texttt{Mean}, and \texttt{Max} as the pooling strategies.
The derived sentence embedding is then input to the word prediction layer and fine-tunes the model to predict the corresponding word.
The word prediction layer is the one that was used for masked language modeling during pre-training.
\citet{DefSent} reported that DefSent performed comparably to SBERT.

\subsection{STS tasks}
We use STS tasks to investigate the properties of sentence embeddings. 
STS tasks evaluate how the semantic similarity between two sentences calculated with a model correlates with a human-labeled similarity score through Pearson and Spearman correlations.
There are two types of settings: supervised and unsupervised.
In the supervised setting, a model learns a regression function that maps a pair of sentences to a similarity score using some of the STS datasets.
In the unsupervised setting, no training is performed on STS datasets, and we compute the similarity between two sentence embeddings, with a similarity score such as cosine similarity.

For the evaluation of the STS tasks, STS12--STS16 \citep{STS12, STS13, STS14, STS15, STS16}, STS Benchmark \citep{STSB}, and SICK-R \citep{SICK} are often used.
Each dataset contains sentence pairs with their semantic similarity scores as gold labels given by real numbers ranging from 0 to 5.
Each of the STS12--STS16 datasets consists of sentence pairs from multiple sources.
For example, STS12 consists of sentence pairs from five sources: \textsl{MSRpar}, \textsl{MSRvid}, \textsl{SMTeuroparl}, \textsl{OnWN}, and \textsl{SMTnews}.
Table \ref{tab:sts-source-statistics} lists the sources of each dataset in STS12--STS16.

\begin{table}[t]
    \centering
    \small
    \tabcolsep 4pt
    \begin{tabular}{@{\ \ }cc@{ }|c|l}
        \hline
        \multicolumn{2}{@{}c@{ }|}{\rule{0pt}{2ex}Sources} & \# & Origin \\
        \hline
        \multirow{5}{*}{STS12}
        & \rule{0pt}{2ex}\textsl{MSRpar} & 750 & newswire \\
        & \textsl{MSRvid} &750 & videos \\
        & \textsl{SMTeuroparl}& 459 & WMT eval. \\
        & \textsl{OnWN} & 750 & glosses \\
        & \textsl{SMTnews} & 399 & WMT eval. \\
        \hline
        \multirow{3}{*}{STS13}
        & \rule{0pt}{2ex}\textsl{FNWN} & 189 & glosses \\
        & \textsl{headlines} & 750 & newswire \\
        & \textsl{OnWN} & 561 & glosses \\
        \hline
        \multirow{6}{*}{STS14}
        & \rule{0pt}{2ex}\textsl{deft-forum} & 450 & forum posts \\
        & \textsl{deft-news} & 300 & news summary\\
        & \textsl{headlines} & 750 & newswire headlines \\
        & \textsl{images} & 750 & image descriptions \\
        & \textsl{OnWN} & 750 & glosses \\
        & \textsl{tweet-news}& 750 & tweet-news pairs \\
        \hline
        \multirow{5}{*}{STS15}
        & \rule{0pt}{2ex}\textsl{answers-forums} & 375 & Q\&A forum answers \\
        & \textsl{answers-students} & 750 & student answers \\
        & \textsl{belief} & 375 & committed belief \\
        & \textsl{headlines} & 750 & newswire headlines \\
        & \textsl{images} & 750 & image descriptions \\
        \hline
        \multirow{5}{*}{STS16}
        & \rule{0pt}{2ex}\textsl{answer-answer} & 254 & Q\&A forum answers \\
        & \textsl{headlines} & 249 & newswire headlines \\
        & \textsl{plagiarism} & 230 & short-answer plag. \\
        & \textsl{postediting} & 244 & MT postedits \\
        & \textsl{question-question} & 209 & Q\&A forum questions \\
        \hline
    \end{tabular}
    \vspace{-0.5ex}
    \caption{Statistics of STS datasets partitioned by source. ``\#'' denotes number of sentence pairs, and ``Origin'' denotes origin of dataset.}
    \label{tab:sts-source-statistics}
    \vspace{-2.5ex}
\end{table}

\subsection{SentEval}
We also compare SBERT and DefSent on SentEval \citep{SentEval} tasks.
SentEval is a widely used toolkit to evaluate the quality of sentence embeddings by measuring the performance on classification tasks.
Since SentEval provides various classification tasks, it is suitable for investigating the properties of sentence embeddings.
SentEval consists of two types of tasks: downstream tasks and probing tasks.
Downstream tasks are binary or multi-class classification tasks, such as sentiment classification in movie reviews and question-type classification.
Probing tasks are classification tasks for linguistic information, such as sentence length and tense classification.

\subsection{Experimental settings}
\label{sec:evaluation-settings}
In the experiments reported in Sections \ref{sec:comparison-of-sentence-embeddings} and \ref{sec:combination-of-sentence-embeddings}, we use BERT-base (bert-base-uncased), BERT-large (bert-large-uncased), RoBERTa-base (roberta-base), and RoBERTa-large (roberta-large) from Transformers \cite{Transformers} as the pre-trained language models and adopt \texttt{Mean} as the pooling strategy.
We use the same settings as \citet{SBERT} and \citet{DefSent} for fine-tuning.
We provide further training details in Appendix \ref{appendix:training-details}, and report the fine-tuning time and computing infrastructure in Appendix \ref{appendix:runtime}.

\section{Comparison of Sentence Embeddings}
\label{sec:comparison-of-sentence-embeddings}
The supervision signal used for fine-tuning sentence embeddings might affect their properties. 
For example, since it is crucial to capture the differences in meaning even when the given sentence pair is superficially similar in the NLI task, SBERT is considered suitable for determining the semantic similarity between superficially similar sentence pairs.
In this section, we attempt to reveal such properties of each type of sentence embeddings.
First, we partition the STS datasets on the basis of the source of the sentence pairs and the superficial similarity of the sentence pair.
We then apply each type of embeddings to the downstream and probing tasks of SentEval.

\begin{figure*}[t!]
    \centering
    \includegraphics[width=0.85\linewidth]{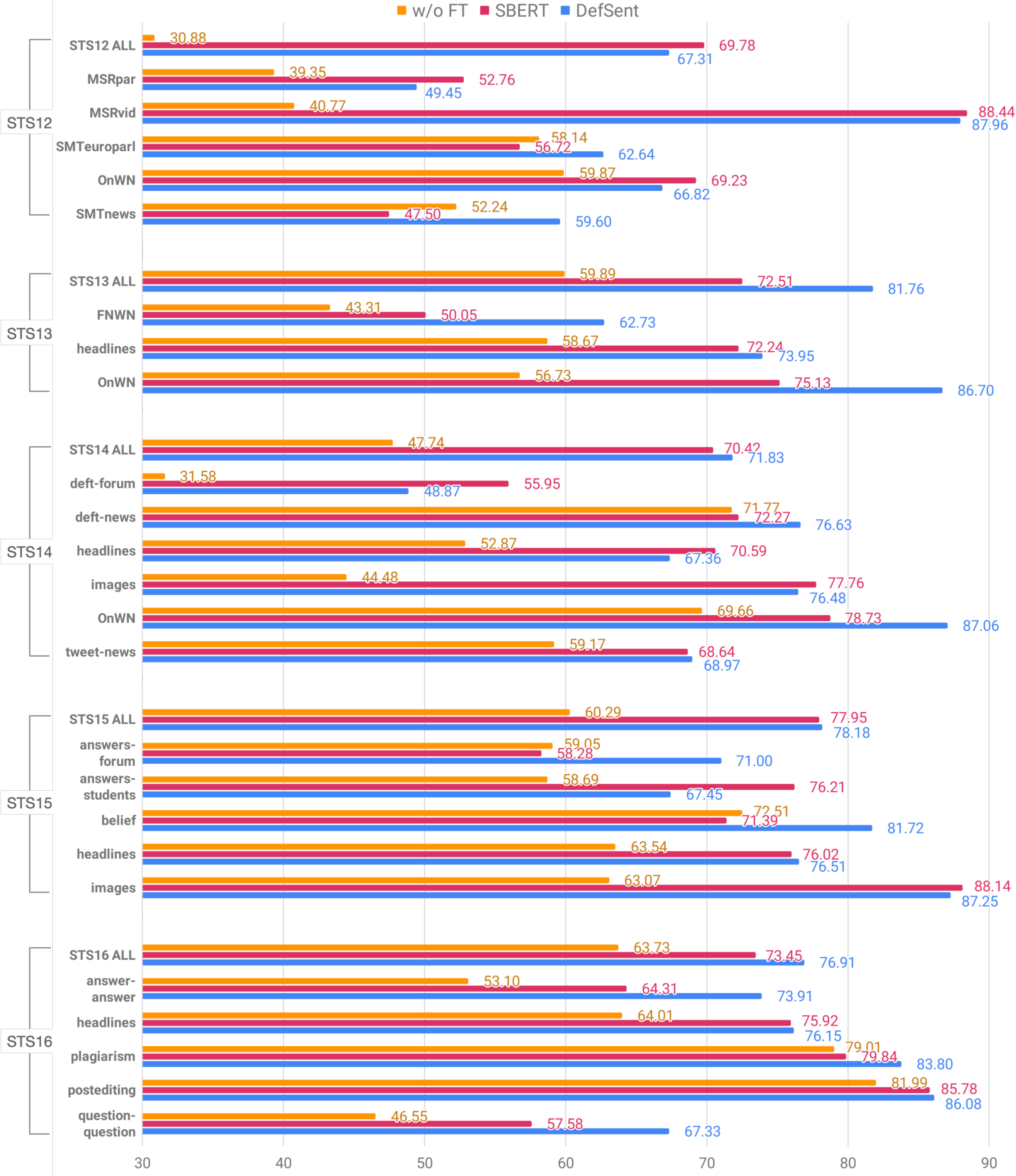}
    \vspace{\FigSpaceTop}
    \caption{
    Spearman's $\rho\times100$ for STS12--STS16 datasets partitioned by source.
    ``STS\# ALL'' denotes the concatenation of all subsets for each STS dataset.
    }
    \label{fig:sts-source}
    \vspace{\FigSpaceBottom}
\end{figure*}

\subsection{STS partitioned by source}
\label{sec:spliting-sts-datasets}
We assume that each sentence embedding method might better capture the meaning of sentences similar to those in the dataset used for fine-tuning, i.e., NLI datasets for SBERT and word dictionaries for DefSent.
Thus, we partition STS12--STS16 datasets in accordance with the source of the sentences and measure the performance for each subset.
We adopt the unsupervised setting.
We calculate Spearman's rank correlation coefficient ($\rho$) between semantic similarity scores and each type of sentence embeddings.
For comparison, we conduct evaluations on the concatenation of all subsets, i.e., the STS datasets without partitioning.
We fine-tune and evaluate SBERT and DefSent 10 times with different seed values and report the average.
We also evaluate the model without fine-tuning (\woft) for comparison.

Figure \ref{fig:sts-source} shows the Spearman's $\rho$ for the subsets of the STS12--STS16 datasets.
It is worth noting that since we use correlations, the evaluation score on the concatenation of all subsets is not the average of the other scores, and in extreme cases it can be smaller than the minimum of the other scores.
We can see that both SBERT and DefSent achieve higher scores than \woft\ on most subsets.
Although DefSent consistently performs better than \woft\ in all subsets, SBERT performs worse than \woft\ in some subsets.
Comparing SBERT and DefSent, when we focus on individual subsets, we can find that there are cases in which SBERT achieves higher scores than DefSent, but we can say that DefSent achieves slightly higher scores as a whole.
DefSent achieves noticeably higher scores than SBERT on \textsl{OnWN} and \textsl{FNWN} of STS13 and \textsl{OnWN} of STS14.
\textsl{OnWN} and \textsl{FNWN} of STS13 are datasets created using definition sentences in OntoNotes, FrameNet, and WordNet.
These results, as expected, indicate that DefSent is capable of adequately representing the meaning of definition sentences.
However, SBERT achieves higher scores than DefSent on \textsl{deft-forum} and \textsl{headlines} of STS14 and \textsl{answer-students} of STS15.
Regarding \textsl{answer-students}, since it is built from a dataset that has a similar format to the NLI datasets \cite{STS15}, it is considered a score such as the one observed is as expected for SBERT, which is trained on the NLI datasets.

\begin{table*}[t]
\centering
\small
\begin{tabular}{@{ \ }ll||cc|ccc}
\hline
\rule{0pt}{2ex}sentence 1 & sentence 2 & Human & Dice & \woft & SBERT & DefSent\\
\hline
\rule{0pt}{2ex}A man is playing a guitar.  & The man is playing the guitar. & 4.909 & 0.800 & 0.906 & 0.985 & 0.978 \\
A man is playing a guitar.  & A guy is playing an instrument. & 3.800 & 0.545 & 0.945 & \textbf{0.646} & 0.895  \\
A man is playing a guitar.  & A man is playing a guitar and singing. & 3.200 & 0.833 & 0.979 & 0.874 & \textbf{0.977} \\
A man is playing a guitar.	& The girl is playing the guitar. & 2.250 & 0.600 & 0.900 & 0.747 & 0.831 \\
A man is playing a guitar.	& A woman is cutting vegetable. & 0.000 & 0.400 & 0.890 & 0.290 & 0.595\\
\hline
\end{tabular}
\vspace{-0.5ex}
\caption{
Example sentence pairs in STS Benchmark datasets and their scores.
``Human'' denotes human-labeled similarity scores, ``Dice'' denotes Dice coefficients, and ``\woft'', ``SBERT'', and ``DefSent'' denote cosine similarities between each sentence embedding computed with BERT without fine-tuning, SBERT, and DefSent, respectively.
The average cosine similarity for \woft\ is 0.816, for SBERT is 0.678, and for DefSent is 0.809.
}
\label{tab:sts-example}
\vspace{-1ex}
\end{table*}

\subsection{STS partitioned by Dice coefficient}
We then explore how the similarity of sentence embeddings is affected by the superficial similarity of the sentences.
Generally speaking, it is considered difficult to correctly order the similarity of a dataset consisting of pairs with high superficial similarity.
However, since the NLI datasets contain a relatively large number of superficially similar sentences, SBERT built on such a dataset is expected to be relatively robust to sentence pairs with high superficial similarity.
To verify whether there is such a tendency, we partition STS Benchmark datasets in accordance with the superficial similarity of the sentences and investigate the performance of each embedding method on the partitioned datasets.
Specifically, we use Dice coefficients between the sets of words in a sentence pair as the superficial similarity, which is defined as
\[
    \mbox{Dice}(S_1, S_2) = \frac{2 |W_1 \cap W_2|}{|W_1|+|W_2|},
\]
where $S_1$ and $S_2$ are the sentence pair, and $W_1$ and $W_2$ are the sets of words in $S1$ and $S2$, respectively.
We sort the sentence pairs in all STS Benchmark datasets including training, development, and test sets in accordance with the Dice coefficient, and partition them into five subsets, that is, grouping 20\% of the sentences from bottom to top.

\begin{figure}[t]
    \centering
    \includegraphics[width=\linewidth]{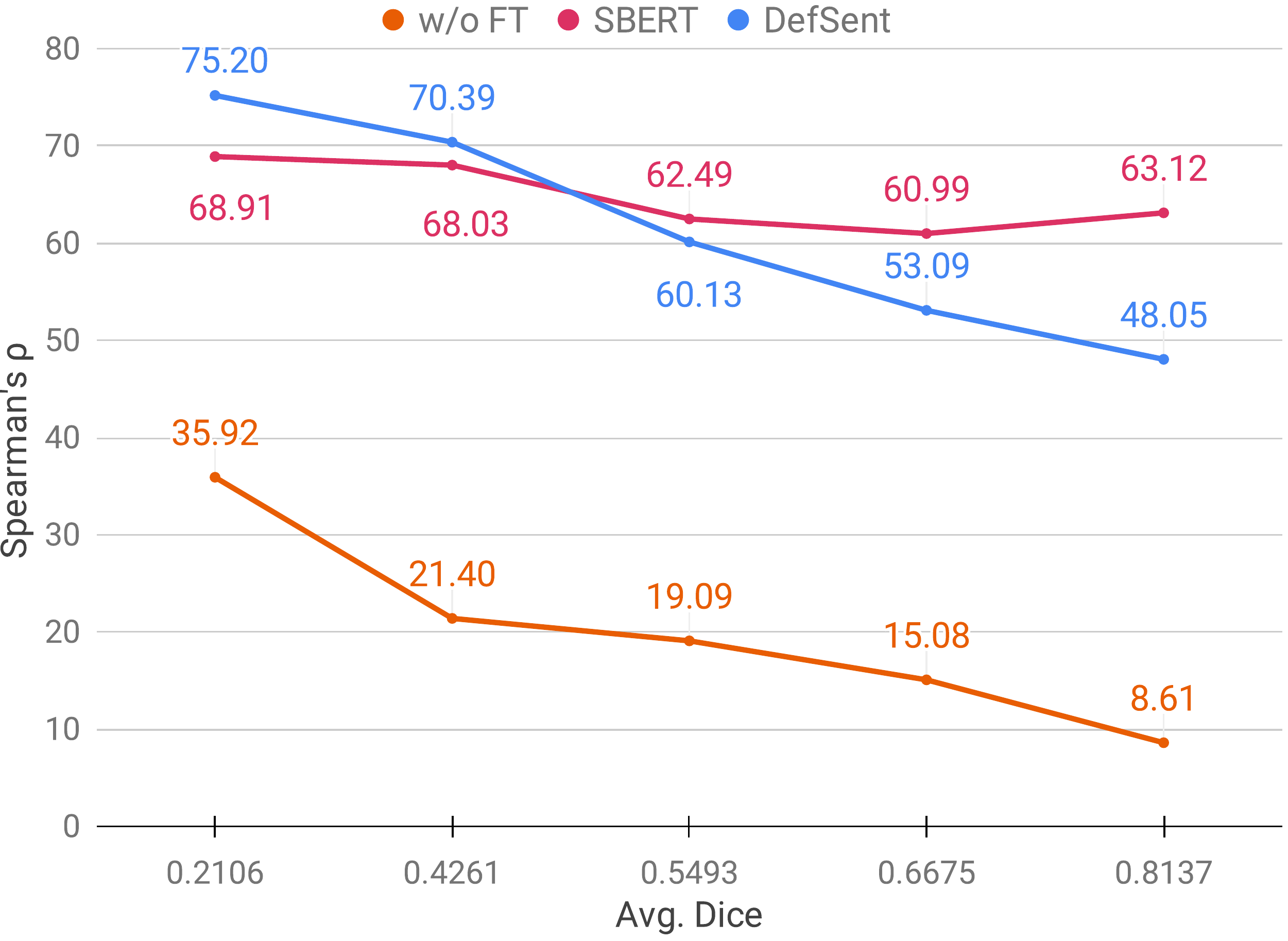}
    \vspace{\FigSpaceTop}
    \caption{
    Spearman’s $\rho \times 100$ for STS Benchmark partitioned in accordance with the ratio of shared words.
    Sentence pairs are more superficially similar to right.
    }
    \label{fig:sts-dice}
    \vspace{\FigSpaceBottom}
\end{figure}

Figure \ref{fig:sts-dice} shows the Spearman's $\rho$ for each subsets.
We can confirm that the subsets with larger Dice coefficients, that is, a higher superficial similarity, tend to be more difficult to rank the semantic similarities.
However, as expected, SBERT is more robust to the subsets with higher superficial similarity, and consequently, SBERT achieves a higher score than DefSent for these subsets, whereas DefSent achieved a higher score than SBERT for the subsets with a lower superficial similarity.

For further investigation, we conduct a qualitative analysis of how superficial similarity affects the behavior of the methods.
Table \ref{tab:sts-example} shows example sentence pairs from STS Benchmark datasets with their human-labeled similarity scores, Dice coefficients, and cosine similarities between each sentence embedding with the respective methods.
As shown in the second row from the top, we observe that each sentence of the pair represents almost the same thing except for minor details (``guitar'' or ``instrument''), but SBERT assigns relatively a much lower similarity than other examples.
As shown in the third row from the top, the similarity score of DefSent is very high, even though the human-labeled score is not that high.
In summary, we can say that SBERT is better at capturing the semantic similarity of superficially similar sentences, while DefSent is better at capturing the similarity of sentences with low superficial similarity.

\subsection{SentEval donwstream tasks}
\label{sec:comparison-senteval-downstream-tasks}
We then apply each type of embeddings to the downstream tasks of SentEval and analyze what type of information each type of embeddings captures that is useful for the downstream task.
We train a logistic regression classifier with 10-fold cross-validation, a batch size of 64, an epoch size of 4, and Adam \citep{Adam} optimizer, the same as the default configurations of SentEval.
Specifically, parameters of sentence embedding models are fixed during training of the classifier.
We fine-tune and evaluate SBERT and DefSent three times with different seed values and report the average of accuracy for each downstream task.
We also evaluate \woft\ for comparison.

\begin{figure}[t]
\centering
\includegraphics[width=\linewidth]{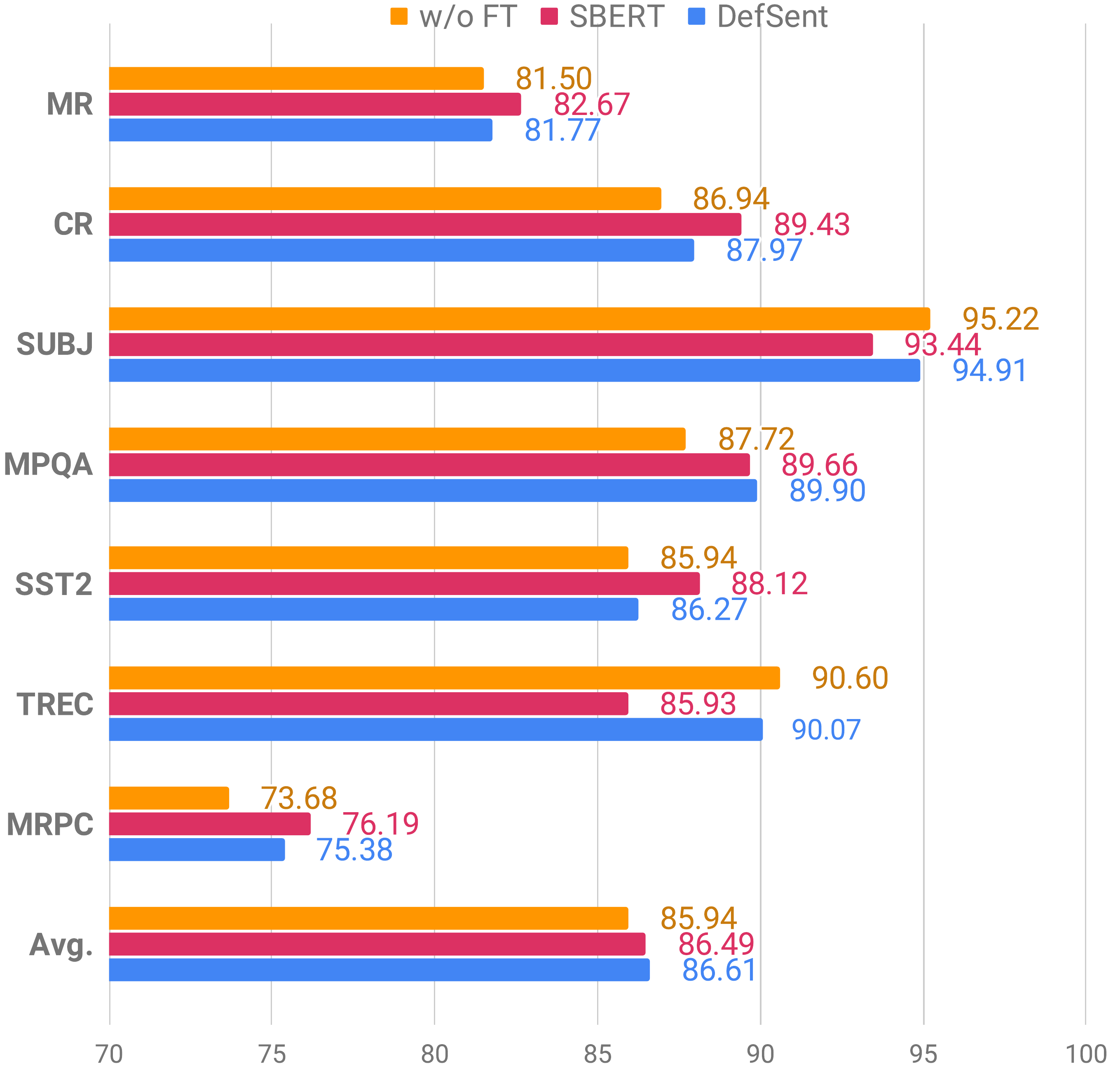}
\vspace{\FigSpaceTop}
\caption{Experimental results on each SentEval \\downstream task with the accuracy (\%).}
\label{fig:senteval-downstream}
\vspace{\FigSpaceBottom}
\end{figure}

Figure \ref{fig:senteval-downstream} shows the accuracy for downstream tasks.
As a whole, SBERT and DefSent perform comparably.
SBERT performs best for MR, CR, SST2, and MRPC.
Since MR, CR, and SST2 are sentiment prediction tasks, it suggests that SBERT encodes the sentiment of sentences into the embedding.
Also, MRPC is a paraphrase-prediction task, which predicts whether two sentences have the same meaning on the basis of their embeddings.
Therefore, MRPC is similar to the NLI task, and thus it is not surprising that SBERT performs better.

\begin{figure}[t]
\centering
\includegraphics[width=\linewidth]{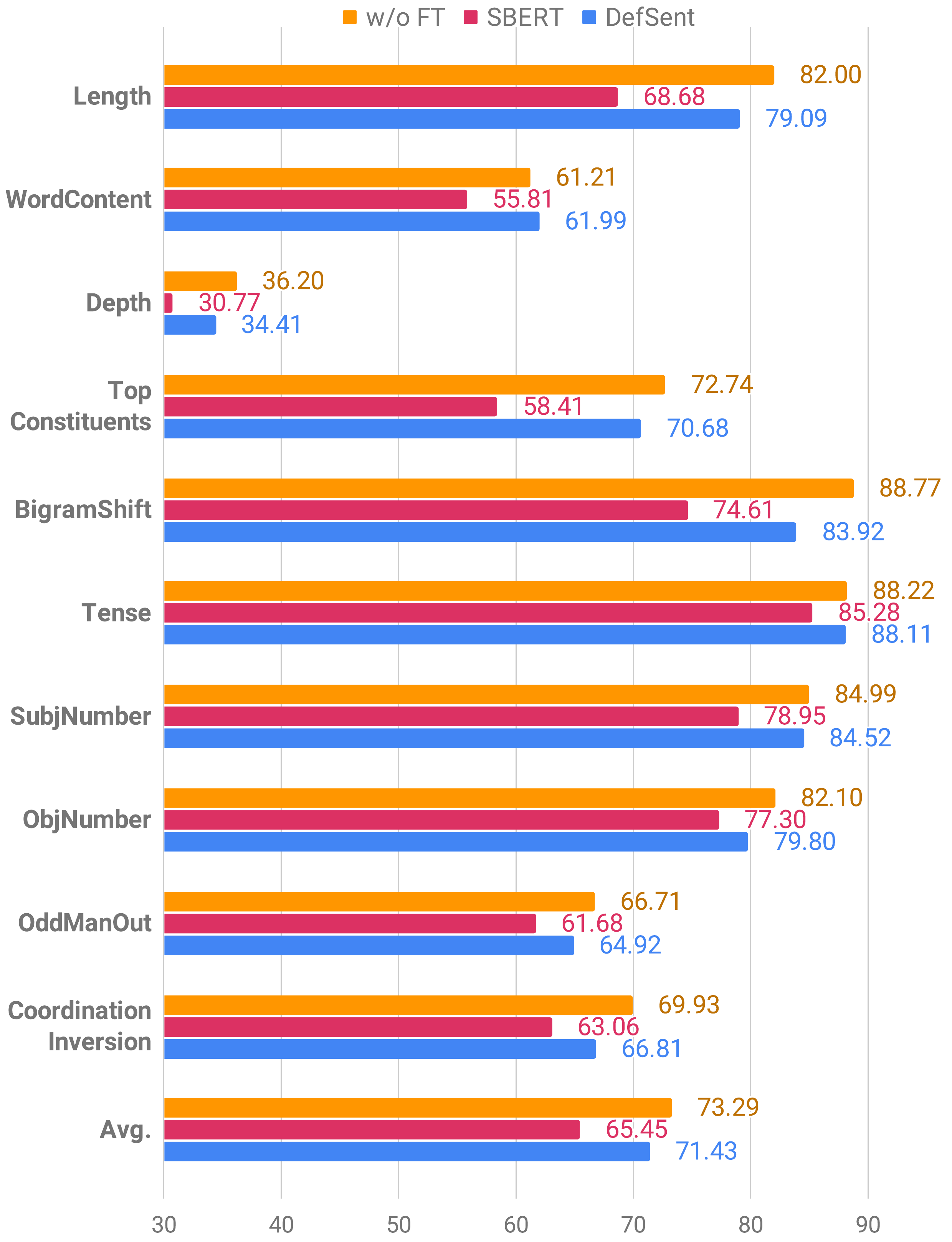}
\vspace{\FigSpaceTop}
\caption{Experimental results on each SentEval \\probing task with the accuracy (\%).}
\label{fig:senteval-probings}
\vspace{\FigSpaceBottom}
\end{figure}

DefSent performs best for MPQA and is comparable to \woft\ for SUBJ and TREC.
MPQA is a phrase-level opinion polarity classification task, and it is necessary to compose the meaning of phrases adequately.
We conjecture that the performance of DefSent is high because DefSent successfully composes the meaning of the corresponding words from the definition sentences during fine-tuning.
It is worth noting that \woft\ performs best for SUBJ and TREC, and SBERT performs much worse for them.
SUBJ is a subjectivity classification task and TREC is a question-type classification task.
Since information about words in sentences is particularly important for these tasks, SBERT is considered to have less information about which words are included in sentences than DefSent and \woft.
Therefore, we can say that SBERT encodes mainly sentiment information into the sentence embedding, and the sentence embedding is suitable for determining whether the meaning is the same.
Also, DefSent successfully composes the meaning of the sentence from its words and encodes information about words the sentence has.

\subsection{SentEval probing tasks}
Finally, we apply each type of embeddings to the probing tasks of SentEval and analyze what type of linguistic information each type of embeddings captures.
We use the same setting as in Section \ref{sec:comparison-senteval-downstream-tasks}.

Figure \ref{fig:senteval-probings} shows the accuracy for probing tasks.
Overall, \woft\ performs best on average, followed by DefSent, and then SBERT.
The overall performance of SBERT is relatively low.
SBERT encodes the semantic information of sentences according to the results of SentEval downstream tasks.
These results also indicate that SBERT encodes semantic information rather than linguistic information such as words in a sentence.
DefSent is comparable to \woft\ in WordContent, Tense, and SubjNumber.
This also indicates that the sentence embeddings from DefSent have information about words the sentence contains.

\begin{table*}[t]
\centering
\small
\begin{tabular}{l|l||c|c|c|c|c|c|c||c}
\hline
\rule{0pt}{2ex}Model & Method & STS12 & STS13 & STS14 & STS15 & STS16 & STS-B & SICK-R & Avg.\\
\hline
\rule{0pt}{2ex}BERT-base & \woft &30.88 & 59.90 & 47.74 & 60.29 & 63.73 & 47.29 & 58.22 & 52.58\\
BERT-base & SBERT &69.78 & 72.51 & 70.42 & 77.95 & 73.45 & 75.96 & 72.26 & 73.19\\
BERT-base & DefSent &67.31 & 81.76 & 71.83 & 78.18 & 76.91 & 76.98 & 73.47 & 75.20\\
BERT-base & \textsc{S+D} &70.71 & \textbf{83.48} & \textbf{76.66} & \textbf{82.00} & \textbf{78.70} & \textbf{80.76} & \textbf{76.83} & \textbf{78.45}\\
BERT-base & \textsc{D+S} & 68.68 & 73.65 & 70.60 & 76.96 & 72.54 & 75.30 & 72.46 & 72.89\\
BERT-base & \textsc{Multi} & 63.10 & 74.34 & 70.30 & 77.64 & 74.08 & 77.35 & 73.42 & 72.89\\
BERT-base & \textsc{Average} & \textbf{72.40} & 81.36 & 75.80 & 81.90 & 77.64 & 79.74 & 75.87 & 77.81\\
BERT-base & \textsc{Concat} & 71.13 & 78.54 & 74.03 & 79.95 & 76.01 & 78.37 & 74.17 & 76.03\\
\hline
\rule{0pt}{2ex}BERT-large & \woft & 27.69 & 55.78 & 44.48 & 51.67 & 61.85 & 47.00 & 53.85 & 48.90\\
BERT-large & SBERT & 70.76 & 73.68 & 72.56 & 79.00 & 74.61 & 77.11 & 72.47 & 74.31\\
BERT-large & DefSent & 63.30 & 82.16 & 72.67 & 79.06 & 77.52 & 77.40 & 74.02 & 75.16\\
BERT-large & \textsc{S+D} & 69.48 & \textbf{83.90} & 76.83 & \textbf{82.61} & \textbf{80.14} & \textbf{81.72} & \textbf{78.77} & \textbf{79.06}\\
BERT-large & \textsc{D+S} & 71.25 & 75.71 & 73.39 & 79.68 & 75.20 & 77.67 & 73.78 & 75.24\\
BERT-large & \textsc{Multi} & 70.33 & 81.16 & 75.84 & 80.02 & 76.52 & 78.65 & 74.30 & 76.69\\
BERT-large & \textsc{Average} & \textbf{71.85} & 82.60 & \textbf{77.33} & 82.52 & 79.12 & 80.71 & 76.30 & 78.63\\
BERT-large & \textsc{Concat} & 71.37 & 80.28 & 76.08 & 81.10 & 77.63 & 79.57 & 74.71 & 77.25\\
\hline
\end{tabular}
\vspace{-1ex}
\caption{Experimental results on unsupervised STS tasks with Spearman's $\rho \times 100$.}
\label{tab:comb-sts}
\vspace{-2.5ex}
\end{table*}

\section{Combination of Sentence Embeddings}
\label{sec:combination-of-sentence-embeddings}
We have shown that SBERT and DefSent have different properties and that they may be complimentary.
This suggests that combining the two methods may yield better sentence embeddings.
Thus, we attempt to combine SBERT and DefSent and evaluate the resulting sentence embeddings on unsupervised STS tasks and SentEval downstream tasks. 
Specifically, we use the following five methods of combining SBERT and DefSent for BERT\footnote{
The experimental results for RoBERT are given in Appendix \ref{appendix:roberta-sts} and \ref{appendix:roberta-senteval}.
}.

\begin{description}
\vspace{-1ex}
  \setlength{\itemsep}{0.25pt}
  \item[\textsc{S+D}] Fine-tuning the pre-trained model with SBERT then with DefSent sequentially.
  \item[\textsc{D+S}] Fine-tuning the pre-trained model with DefSent then with SBERT sequentially.
  \item[\textsc{Multi}] Multi-task learning with SBERT and DefSent. The ratio of the size of the NLI dataset to the dictionary dataset is about 19:1, so we do 19 steps with SBERT and then 1 step with DefSent for the same model.
  \item[\textsc{Average}] Averaging embeddings of separately fine-tuned models with SBERT and DefSent.
  \item[\textsc{Concat}] Concatenate embeddings of separately fine-tuned models with SBERT and DefSent.
\end{description}

\subsection{Evaluation on unsupervised STS tasks}
\label{sec:combination-sts}
We first estimate the resulting sentence embeddings on unsupervised STS tasks.
We use the same settings described in Section \ref{sec:evaluation-settings}.
We use STS12--STS16, STS Benchmark test set (STS-B), and SICK-Relatedness (SICK-R) for the evaluation.
We compute sentence similarities by using the cosine similarity of sentence embeddings derived from the respective combinations and calculate Spearman’s $\rho$ with gold labels.
We conduct fine-tuning and evaluations 10 times with different seed values and report the average.

Table \ref{tab:comb-sts} shows the experimental results.
The combinations \textsc{S+D}, \textsc{Average}, and \textsc{Concat} always outperform SBERT and DefSent.
Among them, \textsc{S+D} achieves the best average score for base and large models.
However we cannot confirm much performance improvement with \textsc{D+S} and \textsc{Multi}.
We leave an analysis of what affects this difference in performances as future work.

\begin{table*}[t]
\centering
\small
\begin{tabular}{@{ \ }l@{ \ }||l|l|c|c|c|c|c|c||c}
\hline
\rule{0pt}{2ex}Model & Method & MR & CR & SUBJ & MPQA & SST-2 & TREC & MRPC & Avg. \\
\hline
\rule{0pt}{2ex}BERT-base & \woft & 81.50 & 86.94 & \textbf{95.22} & 87.72 & 85.94 & 90.60 & 73.68 & 85.94\\
BERT-base & SBERT & 82.67 & 89.43 & 93.44 & 89.66 & 88.12 & 85.93 & 76.19 & 86.49\\
BERT-base & DefSent & 81.77 & 87.97 & 94.91 & 89.90 & 86.27 & 90.07 & 75.38 & 86.61\\
BERT-base & \textsc{S+D} & 81.29 & 89.10 & 93.99 & 90.09 & 86.69 & 89.33 & 77.08 & 86.80\\
BERT-base & \textsc{D+S} & 82.43 & 89.22 & 93.24 & 90.16 & \textbf{88.98} & 83.33 & 75.27 & 86.09\\
BERT-base & \textsc{Multi} & 81.73 & 88.80 & 93.17 & 89.27 & 87.28 & 87.87 & 75.54 & 86.23\\
BERT-base & \textsc{Average} & 83.17 & 89.50 & 94.67 & 90.35 & 88.50 & 89.67 & 76.41 & 87.47\\
BERT-base & \textsc{Concat} & \textbf{83.24} & \textbf{89.64} & 95.18 & \textbf{90.51} & 88.94 & \textbf{90.60} & \textbf{77.37} & \textbf{87.93}\\
\hline
\rule{0pt}{2ex}BERT-large & \woft & 84.30 & 89.16 & \textbf{95.60} & 86.65 & 89.29 & \textbf{91.40} & 71.65 & 86.86\\
BERT-large & SBERT & 84.76 & 90.61 & 94.08 & 90.04 & 90.77 & 85.47 & 75.90 & 87.38\\
BERT-large & DefSent & 84.54 & 89.40 & 95.55 & 90.04 & 89.49 & 88.73 & 74.82 & 87.51\\
BERT-large & \textsc{S+D} & 84.01 & 90.49 & 95.07 & 90.50 & 90.35 & 90.20 & 75.61 & 88.03\\
BERT-large & \textsc{D+S} & 84.55 & 90.68 & 93.46 & 90.22 & 90.21 & 84.73 & 75.01 & 86.98 \\
BERT-large & \textsc{Multi} & 84.63 & 90.56 & 94.10 & 89.85 & 90.23 & 88.70 & 76.56 & 87.80\\
BERT-large & \textsc{Average} & 85.46 & \textbf{90.92} & 95.20 & 90.53 & 91.27 & 88.27 & \textbf{77.00} & 88.38\\
BERT-large & \textsc{Concat} & \textbf{85.53} & 90.83 & 95.27 & \textbf{90.66} & \textbf{91.95} & 89.60 & 75.88 & \textbf{88.53}\\
\hline
\end{tabular}
\vspace{-1ex}
\caption{Experimental results on each SentEval task with the accuracy (\%).}
\label{tab:comb-senteval}
\vspace{-2.5ex}
\end{table*}

\subsection{Evaluation on the SentEval tasks}
\label{sec:combination-senteval}
We then estimate the resulting sentence embeddings on the SentEval tasks.
We use the same settings described in Section \ref{sec:comparison-senteval-downstream-tasks}.
We conduct fine-tuning and evaluations three times with different seed values and report the average.

Table \ref{tab:comb-senteval} shows the results.
We can see that \textsc{Concat} achieves the highest average score but it should be noted that since SentEval performed supervised learning of a logistic regression classifier, the high dimensionality of the sentence embeddings of \textsc{Concat} is advantageous.
Other than \textsc{Concat}, \textsc{Average} performs relatively well, which always outperforms \textsc{S+D}, \textsc{D+S}, and \textsc{Multi}, unlike in the STS tasks.
This suggests that fine-tuning the same model with different tasks might degrade the generalization ability.

\section{Related work}

Sentence embedding has been studied intensively.
\citet{SkipThought} proposed SkipThought, which trains a sentence embedding model by predicting the previous and next sentence from the embedding of a given sentence.
\citet{InferSent} proposed InferSent, which trains a sentence embedding model built on BiLSTM in a Siamese network architecture on the NLI task.
\citet{USE} proposed Universal Sentence Encoder (USE), which is trained on an NLI dataset, and has also shown the effectiveness of NLI datasets in obtaining sophisticated sentence embeddings.

Recently, methods that leverage pre-trained language models to acquire sentence embeddings have attracted much attention.
Pre-trained language models, such as BERT \citep{BERT} and RoBERTa \citep{RoBERTa}, acquire linguistic knowledge by training on large texts and perform well on downstream tasks.
Pre-trained models are also considered helpful for sentence embedding.
There are two types of methods based on pre-trained models: unsupervised and supervised.

Unsupervised methods do not require labeled text but exploit the properties of pre-trained language models or create training data artificially.
\citet{BERT-flow} showed that the sentence embedding space of BERT is anisotropic, and proposed BERT-flow, which learns a map to an isotropic Gaussian distribution to obtain sentence embedding.
Several studies have also been based on contrastive learning, and are different in the way to make positive examples:
DeCLUTR \citep{DeCLUTR} takes into account different spans of the same document as positives;
ConSERT \citep{ConSERT} takes into account a pair of an original sentence and a collapsed sentence as positives;
unsupervised SimCSE \citep{SimCSE} takes into account the corresponding embeddings of the same sentence with different dropout masks applied as positives.

Supervised methods use labeled text to encode higher-level semantic information.
Supervised methods generally produce more sophisticated sentence embeddings than unsupervised methods.
In addition to SBERT and DefSent, supervised SimCSE \citep{SimCSE} is one of the supervised sentence embedding methods.
Supervised SimCSE fine-tunes BERT by contrastive learning using entailment pairs in the NLI datasets as positives.

\section{Conclusion}
In this paper, we empirically investigated the influence of supervision signals used for obtaining sentence embeddings.
We focused on two methods: SBERT, which uses NLI datasets, and DefSent, which uses word dictionaries.
We showed that there is a difference in the ability to order the similarity of sentences depending on their source or superficial similarity by comparing their performances on subsets of the STS datasets and tasks of SentEval.
We found that SBERT is suitable for superficially similar sentence pairs because SBERT is based on the NLI datasets that contain a relatively large number of superficially similar sentences, whereas DefSent is suitable for sentence pairs that need to represent the compositional meaning because DefSent is based on definition sentences of a dictionary.

We also showed that SBERT performed better in tasks where sentiment information was important, while DefSent performed better in tasks where information about words and the compositionality of meaning were important by comparing their performances on downstream and probing tasks of SentEval.
Finally, we demonstrated that combining the two methods yielded substantially better performance than the respective methods on unsupervised STS tasks and downstream tasks of SentEval.

For future work, we will expand the scope of our analysis to other pre-trained language models and sentence embedding methods to obtain insights for better sentence embeddings.
In addition, We will investigate how those combination methods affect the properties of resulting sentence embeddings and 
explore how to effectively combine unsupervised sentence embedding methods, which have recently achieved good performance, such as DeCLUTR \citep{DeCLUTR} and unsupervised SimCSE \citep{SimCSE}, with supervised sentenece embedding methods.
Moreover, the combination of unsupervised methods, which have recently achieved good performance, such as DeCLUTR \citep{DeCLUTR} and unsupervised SimCSE \citep{SimCSE}, and supervised methods should also be promising.

\section*{Acknowledgements}
This work was supported by JSPS KAKENHI Grant Number 21H04901.


\bibliography{anthology,custom}
\bibliographystyle{acl_natbib}

\clearpage
\appendix

\section{Training Details}
\label{appendix:training-details}
For fine-tuning of SBERT and DefSent, we use a batch size of 16, an epoch size of 1, Adam \citep{Adam} optimizer with $\beta_1=0.9$, $\beta_2=0.999$, and a linear learning rate warm-up over 10\% of training steps for each, as the same setting as \citet{SBERT} and \citet{DefSent}.
We choose the learning rate that achieves the highest average score on the validation set for each respective model by fine-tuning three times with different seed values at each learning rate in a range of $x\times 10^{-6}, x\in\{1, 2, 5, 10, 20, 50\}$.
We also use smart batching, and the max sequence length is 128 for training efficiency.

\section{Average Runtime and Computing Infrastructure}
\label{appendix:runtime}
Fine-tuning of SBERT with BERT-base and RoBERTa-base took about 120 minutes on a single NVIDIA GeForce GTX 1080 Ti.
Fine-tuning of DefSent with BERT-base and RoBERTa-base took about 10 minutes on a single NVIDIA GeForce GTX 1080 Ti.
Fine-tuning of SBERT with BERT-large and RoBERTa-large took about 130 minutes on a single Quadro GV100.
Fine-tuning of DefSent with BERT-large and RoBERTa-large took about 15 minutes on a single Quadro GV100.

\section{The details of evaluation on unsupervised STS tasks of RoBERTa}
\label{appendix:roberta-sts}
Table \ref{tab:comb-roberta-sts} shows the average of Spearman's $rho$ for RoBERTa-base and RoBERTa-large on unsupervised STS tasks.

\section{The details of evaluation on SentEval of RoBERTa}
\label{appendix:roberta-senteval}
Table \ref{tab:comb-roberta-senteval} shows the average of accuracy for RoBERTa-base and RoBERTa-large on SentEval.

\begin{table*}[t]
\centering
\small
\begin{tabular}{@{ \ }l@{ \ }|l||c|c|c|c|c|c|c||c}
\hline
\rule{0pt}{2ex}Model & Method & STS12 & STS13 & STS14 & STS15 & STS16 & STS-B & SICK-R  & Avg.\\
\hline
\rule{0pt}{2ex}RoBERTa-base & \woft & 30.61& 55.55& 46.78& 58.43 & 61.21 & 54.36 & 62.17 & 52.73\\
RoBERTa-base & SBERT & 70.20 & 74.44 & 71.86 & 78.70 & 74.47 & 76.92 & 72.11 & 74.10\\
RoBERTa-base & DefSent &60.05 & 76.16 & 69.06 & 74.07 & 77.86 & 76.58 & 74.05 & 72.55\\
RoBERTa-base & \textsc{S+D} & 73.19 & 83.86 & 77.45 & 83.32 & 78.88 & 80.67 & 76.97 & 79.19\\
RoBERTa-base & \textsc{D+S} & 70.97 & 75.07 & 72.50 & 79.04 & 74.56 & 77.13 & 72.81 & 74.58\\
RoBERTa-base & \textsc{Multi} & 69.27 & 77.34 & 73.10 & 80.68 & 76.08 & 77.97 & 73.61 & 75.44\\
RoBERTa-base & \textsc{Average} & 71.61 & 78.65 & 74.65 & 80.30 & 76.71 & 78.56 & 74.04 & 76.36\\
RoBERTa-base & \textsc{Concat} &70.69 & 76.03 & 72.92 & 79.08 & 75.34 & 77.50 & 72.73 & 74.90\\
\hline
\rule{0pt}{2ex}RoBERTa-large& \woft &26.00 & 54.35 & 44.10 & 56.35 & 60.37 & 47.01 & 58.11 &49.47\\
RoBERTa-large & SBERT & 74.04 & 79.47 & 75.47 & 82.77 & 79.50 & 80.49 & 74.19 & 77.99\\
RoBERTa-large & DefSent & 57.79 & 74.67 & 69.01 & 72.98 & 75.48 & 77.39 & 72.55 & 71.41\\
RoBERTa-large & \textsc{S+D} & 66.62 & 79.60 & 75.81 & 77.91 & 78.45 & 80.46 & 77.45 & 76.61\\
RoBERTa-large & \textsc{D+S} & 74.18 & 79.81 & 76.38 & 82.85 & 78.78 & 80.38 & 74.86 & 78.18\\
RoBERTa-large & \textsc{Multi} & 61.34 & 57.43 & 60.17 & 75.56 & 73.78 & 74.92 & 70.10 & 67.62\\
RoBERTa-large & \textsc{Average} & 73.43 & 82.97 & 77.85 & 83.82 & 80.65 & 82.09 & 75.91 & 79.53\\
RoBERTa-large & \textsc{Concat} & 74.04 & 80.96 & 76.60 & 83.20 & 80.33 & 81.24 & 74.77 & 78.73\\
\hline
\end{tabular}
\caption{Experimental results on unsupervised STS tasks with Spearman's $\rho \times 100$.}
\label{tab:comb-roberta-sts}
\end{table*}

\begin{table*}[t]
\centering
\small
\begin{tabular}{@{ \ }l@{ \ }||l|l|c|c|c|c|c|c||c}
\hline
\rule{0pt}{2ex}Model & Method & MR & CR & SUBJ & MPQA & SST-2 & TREC & MRPC & Avg. \\
\hline
\rule{0pt}{2ex}RoBERTa-base & \woft & 84.35 & 88.19 & 95.28 & 86.49 & 89.46 & 93.20 & 74.20 & 87.31\\
RoBERTa-base & SBERT & 85.35 & 91.50 & 93.15 & 90.95 & 92.06 & 87.07 & 76.62 & 88.10\\
RoBERTa-base & DefSent & 84.70 & 91.15 & 94.55 & 90.56 & 89.88 & 92.40 & 76.43 & 88.52\\
RoBERTa-base & \textsc{S+D} & 85.04 & 91.40 & 94.17 & 90.81 & 90.63 & 92.00 & 77.14 & 88.74\\
RoBERTa-base & \textsc{D+S} & 85.20 & 91.34 & 93.45 & 90.84 & 92.20 & 88.20 & 76.29 & 88.22\\
RoBERTa-base & \textsc{Multi} & 85.15 & 91.00 & 93.25 & 90.69 & 91.47 & 89.67 & 77.08 & 88.33\\
RoBERTa-base & \textsc{Average}& 85.57 & 91.66 & 94.01 & 91.14 & 92.55 & 89.67 & 78.12 & 88.96\\
RoBERTa-base & \textsc{Concat}& 86.04 & 91.68 & 94.70 & 91.02 & 92.40 & 93.93 & 78.24 & 89.72\\
\hline
\rule{0pt}{2ex}RoBERTa-large& \woft & 85.46 & 88.72 & 96.04 & 88.34 & 91.27 & 93.80 & 73.80 & 88.20\\
RoBERTa-large & SBERT & 87.35 & 92.56 & 94.13 & 90.99 & 92.77 & 92.20 & 76.00 & 89.43\\
RoBERTa-large & DefSent & 86.28 & 91.14 & 95.12 & 90.97 & 90.74 & 92.33 & 73.74 & 88.62\\
RoBERTa-large & \textsc{S+D} & 86.77 & 92.28 & 94.68 & 91.22 & 91.98 & 92.60 & 77.51 & 89.58\\
RoBERTa-large & \textsc{D+S} & 87.02 & 92.40 & 93.62 & 90.80 & 92.59 & 90.93 & 77.35 & 89.25\\
RoBERTa-large & \textsc{Multi}& 87.52 & 92.56 & 94.39 & 91.09 & 93.15 & 91.60 & 76.69 & 89.57\\
RoBERTa-large & \textsc{Average}& 87.82 & 92.81 & 94.69 & 91.36 & 93.24 & 93.93 & 77.49 & 90.19\\
RoBERTa-large & \textsc{Concat} & 87.87 & 92.84 & 95.22 & 91.64 & 93.06 & 94.27 & 76.23 & 90.16\\
\hline
\end{tabular}
\caption{Experimental results on each SentEval task with the accuracy (\%).}
\label{tab:comb-roberta-senteval}
\end{table*}

\end{document}